# Enhanced UAV Path Planning Using the Tangent Intersection Guidance (TIG) Algorithm

*Hichem Cheriet, Khellat Kihel Badra, Chouraqui Samira*

**Abstract:**
*Efficient and safe navigation of Unmanned Aerial Vehicles (UAVs) is critical for various applications, including combat support, package delivery and Search and Rescue Operations. This paper introduces the Tangent Intersection Guidance (TIG) algorithm, an advanced approach for UAV path planning in both static and dynamic environments. The algorithm uses the elliptic tangent intersection method to generate feasible paths. It generates two sub-paths for each threat, selects the optimal route based on a heuristic rule, and iteratively refines the path until the target is reached. Considering the UAV kinematic and dynamic constraints, a modified smoothing technique based on quadratic Bézier curves is adopted to generate a smooth and efficient route. Experimental results show that the TIG algorithm can generate the shortest path in less time, starting from 0.01 seconds, with fewer turning angles compared to A\*, PRM, RRT\*, Tangent Graph, and Static APPATT algorithms in static environments. Furthermore, in completely unknown and partially known environments, TIG demonstrates efficient real-time path planning capabilities for collision avoidance, outperforming APF and Dynamic APPATT algorithms.*

## 1. Introduction

In recent years, Unmanned Aerial Vehicles (UAVs) have significantly transformed many fields, such as cooperative combat [1], surveillance and security [2], disaster rescue [3], [4], package delivery [5] [6], traffic inspection [7] and target reconnaissance [8]. However, the path planning problem remains a challenge, hindering UAVs' ability to navigate through complex environments cluttered with threats while considering several factors such as obstacle avoidance, trajectory feasibility, energy consumption, and path length. Addressing this challenge is crucial to facilitate the effective application of UAVs and increase mission success rates. Path planning can be divided into two main approaches: static path planning and dynamic path planning. Static path planning, also known as global path planning, involves determining a path for the UAV from a start point to a destination point before the mission begins. This approach typically relies on a predefined map of the environment and considers static obstacles. On the other hand, dynamic path planning, also known as local path planning, involves adapting the UAV's trajectory in real-time based on changing environmental conditions and obstacles encountered during the mission. The path calculation is performed onboard the UAV itself, allowing it to react swiftly to unexpected threats or alterations in the environment. This approach requires continuous sensing and decision-making capabilities to navigate safely and efficiently through the changing environments. Dynamic path planning algorithms are crucial for missions where the environment is uncertain or subject to frequent changes, enabling the UAV to autonomously adjust its path to achieve its objectives while avoiding collisions and optimizing performance. However, it is more computationally expensive, and the planned path may not necessarily be the most efficient. In static path planning, the A* algorithm is known for its effectiveness in finding an optimal path to a target while avoiding obstacles. However, in scenarios involving UAVs, the generated paths may not always be suitable. This is due to the unique constraints faced by UAVs, such as limited maneuverability and altitude restrictions. Also, the existing tangent graph methods need to handle the entire map, which can be time-consuming and poor solution quality, especially in complex environments. As a result, to efficiently plan collision-free paths in static and dynamic environments, this paper proposes a novel autonomous path planning algorithm based on the Tangent Intersection strategy. This algorithm improves the classical Tangent graph-based method and gives smooth suitability for UAVs.

The primary contributions of this paper can be summarized as follows:

1) We introduce a novel path-planning algorithm named Tangent Intersection Guidance (TIG), which is based on the tangent graph. Initially, the algorithm creates tangent lines for all obstacles in the environment. Subsequently, a heuristic rule is applied to select the best sub-path. Then, the algorithm is iteratively repeated until the target point is reached. This approach significantly reduces path length and gives a higher smoothness.

2) Extensive experiments were conducted in static and dynamic environments using random dense maps with elliptic obstacles to validate the effectiveness of the proposed Path Planner. Simulation results demonstrate its capability to efficiently plan high-quality collision-free paths, particularly in dense environments.

The rest of this paper is organized as follows. Section 2 reviews related works in the field. Section 3 defines the path planning problem. Section 4 provides a detailed presentation of the proposed algorithm. Com-putational and comparative results are discussed in Section 5, and the conclusion is presented in Section 6.

## 2. Related Works

Path planning is one of the fundamental tasks for UAVs. It can be divided into five approaches, namely graph-based approaches, sampling-based methods, potential field methods, meta-heuristic methods, and machine learning techniques [9]. Graph-based methods utilize graph representations of the environment to model feasible paths. They often employ algorithms like A* [10], Dijkstra [11], Voronoi diagram [12], the tangent graph, and so on. The A* algorithm is a commonly used method in path planning that efficiently finds the shortest path. Voronoi's diagram's main idea is to partition the environment into regions based on proximity to obstacles. It can be used to generate paths by connecting the centroids of Voronoi cells. The generated Voronoi paths are far from obstacles, which ensures safety, whereas the planned path is not guaranteed to be optimal. Dijkstra's algorithm can find the shortest path, but it requires more space to store the nodes in dense graphs. These three methods are usually employed in a known static environment and cannot be used to perform UAV path planning in a dynamic environment [13].

Robert [14] proposes an algorithm based on tangent graphs, using common tangents of polygons to determine a feasible path. However, this method proves to be computationally expensive, particularly in high-dimensional environments. Furthermore, it fails to ensure a safe distance between the robot and the obstacles. Chen et al. [15] improve the tangent graph algorithm by enclosing obstacles with circular shapes. However, this approach wastes open space areas, resulting in longer path lengths. Liu et al. [16] enclosed the obstacles in ellipses to address the limitation of circular obstacles. This strategy reduces wasted areas and can result in shorter paths. However, despite this improvement, the path accuracy for UAVs in real-life scenarios remains insufficient. An improved version of the A* algorithm called BAA* was proposed by Wu et al. [17] based on multi-direction. This algorithm outperforms traditional A* in terms of both path length and execution time. Nevertheless, it requires parameter tuning and does not account for low-scale UAVs. Yuan et al. [18] propose an improved lazy theta* algorithm for UAV path planning, utilizing an octree map to reduce search nodes and adjust heuristic weights for precision and speed. Simulation and real-flight tests confirm its effectiveness in complex environments with multiple constraints. Bazeela et al. [19] propose an improved tangent graph intersection algorithm called ATGP-TI, which employs a tangent intersection technique and heuristics to find optimal paths for unmanned aerial vehicles. However, the algorithm is computationally expensive, especially in high-dimensional spaces. In conclusion, the current methods relying on tangent graphs still encounter computational challenges due to the necessity of constructing tangents for the entire map, and the quality of the planned path remains inadequate for unmanned aerial vehicles.

Visibility graph-based methods have also been explored for UAV path planning due to their ability to generate optimal paths in structured environments. Liu et al. [20] introduced a tangent graph approach for mobile robots, effectively handling polygonal and curved obstacles. Blasi et al. [21] extended visibility graphs to UAV path planning in 3D constrained environments using layered essential visibility graphs, while Shah and Gupta [22] focused on accelerating A* search on visibility graphs over quadtrees for long-range path planning. However, visibility graph methods require preprocessing of all obstacle edges and can become computationally expensive in environments with numerous obstacles. Huan et al. [23] propose an improved tangent graph algorithm called AP-PATT. The algorithm demonstrates improved results regarding path quality and computational complexity (0.05 seconds). However, the generated path may be infeasible in some situations (Fig.3). The authors also applied a B-spline curve to smooth the generated path, which may lead to collisions(Fig.9). Cheriet et al. [24] propose an enhanced version of the tangent algorithm based on the A* method, called the Tangent A* planner, which uses elliptical obstacles. However, it can be computationally intensive and is limited by the ellipse approximations. Moreover, these visibility and tangent graphs often generate paths that pass too close to obstacles, necessitating additional safety navigation measures.

Potential field algorithms, such as artificial potential field (APF) [25] and vector field histogram (VFH) [26], have been extensively applied in the field of path planning, particularly in dynamic environments. While these algorithms are known for their ability to generate smooth trajectories, they can also potentially become trapped in local minima.

Sampling-based methods such as the Probabilistic Roadmap Planner (PRM) and Rapidly Exploring Random Trees (RRT) are particularly effective in high-dimensional spaces, which are common in UAV path planning. PRM algorithm succeeds at handling complex environments with obstacles, narrow passages, and dynamic changes. However, building and storing the roadmap can be computationally expensive, especially in high-dimensional spaces, and may require significant memory [27]. On the other hand, the RRT algorithm does not necessitate sampling the entire space and constructing the roadmap before the mission, thereby reducing computational costs. RRT also operates efficiently in complex environments. Nevertheless, the quality of the path remains challenging for UAVs [28]. In their research [29], Zhou et al. propose a Depth Sorting Fast Search (DSFS) algorithm to enhance path planning efficiency for underwater gravity-aided navigation. This new method improves the Quick Rapidly-exploring Random Trees* (Q-RRT*) algorithm [30], and the comparative experiments show that DSFS improves computational effi-ciency over Q-RRT*.

Artificial intelligence methods such as genetic algorithms (GA) [31], particle swarm optimization (PSO) [32], ant colony optimization (ACO) [33], and Grey Wolf Optimization (GWO) [34]..etc. Researchers have also proposed variants of these optimization algorithms for addressing UAV path planning problems, particularly in complex environments. However, these methods can become computationally time-consuming, especially when dealing with highly dense environments, and they do not always guarantee an optimal solution.

Also, machine learning methods such as support vector machines [35], neural networks [36], and deep reinforcement learning [37] have been applied to address path-planning challenges. One advantage of these methods is their ability to learn from input data and adapt to the mission environment. However, the training data and the computational resources required for training can be costly and require a lot of time. In addition, many research papers have been proposed. For example, Artificial Neural Networks using Radial Basis Functions (RBF-ANN) [38], improved Deep Q-Network (DQN) [39], and Opportunistic Hamilton-Jacobi-Bellman (oHJB) [40].

The aforementioned survey shows that graph-based methods are computationally demanding, particularly in complex environments, because they need to build the graphs for the entire map. Additionally, it is difficult for the UAV to balance between the planned path quality and the algorithm time.

To address the limitations of traditional tangent graph-based and visibility graph-based approaches, we propose the Tangent Intersection Guidance (TIG) algorithm, a novel tangent-graph path planning method for UAVs. The main contributions of our work are:

- Efficient Graph Construction: Unlike visibility graph-based methods that require precomputing all visibility edges, TIG calculates only the tangents lines of obstacles based on the heuristic function, significantly reducing computational complexity in large environments.

- Collision Detection: Traditional tangent graph methods rely on the line-of-sight algorithm for collision detection, which leads to higher computational costs, especially in the presence of numerous obstacles. TIG instead relies solely on the line-ellipse intersection equation to detect collisions, improving efficiency.

- Adaptability in Dynamic Environments: Visibility graphs and tangent graphs typically require full graph recomputation when obstacles change. TIG efficiently adapts to environmental changes by dividing the environment into smaller sub-environments, enabling real-time path adjustments.

- Safe Navigation: Unlike existing methods, TIG incorporates a safety margin between the planned path and obstacles, ensuring robust collision avoidance and enhancing the reliability of UAV navigation in cluttered environments.

- Effective Waypoint Generation: While tangent-graph methods relying on tangent intersection to create waypoints, the TIG algorithm employs an optimized waypoint generation technique that ensures a smoother trajectory, reducing sharp turns and improving the overall flight efficiency of UAVs.

- Computational Performance: Compared to existing approaches, TIG achieves faster path planning while ensuring collision-free and optimal trajectories, making it highly suitable for UAV navigation in complex environments.

## 3. Problem statement

As shown in Fig.1, a drone delivering goods in a low-altitude complex environment cluttered with obstacles, such as buildings, trees, and bridges from a start point S to a target point T. The drone encounters obstacle avoidance challenges in this particular environment, making it necessary to plan a feasible path without collisions between these points. The three-dimensional scenario can be represented in a two-dimensional scenario to reduce algorithm running time.

The shapes of obstacles typically represented as polygons (e.g., buildings with sharp corners or irregular shapes) are not uniform, requiring higher computational resources and leading to unsmooth paths for the UAV. Additionally, the planned paths using these shapes are closer to obstacles, which increases the collision rate. To address these challenges, the paper represents obstacles as ellipses rather than exact polygonal representations, ensuring a smooth and efficient path. Ellipses provide a more manageable mathematical form for path planning, allowing the use of tangent-based methods that can quickly compute safe paths while maintaining a smooth trajectory. Additionally, by adding a safety distance into the elliptical representation, the UAV maintains a safety zone around obstacles, which reduces the risk of collision. This safety distance can be adjusted depending on the UAV size and the specific mission requirements. The obstacles can be represented as:

$$\frac{((x-x_k)\cos\theta+(y-y_k)\sin\theta)^2}{(a+r_{\text{safe}})^2}+ \frac{(-(x-x_k)\sin\theta+(y-y_k)\cos\theta)^2}{(b+r_{\text{safe}})^2}=1 \quad (1)$$

where $x$ and $y$ are the coordinates of any point on the ellipse, $x_k$ and $y_k$ are the coordinates of the center of the ellipse, $a$ is the semi-major axis, $b$ is the semi-minor axis, $r_{\text{safe}}$ is the safety distance added to both the semi-major and semi-minor axes, and $\theta$ is the rotation angle of the ellipse measured counterclockwise from the positive $x$-axis.

In the previous scenario, there are N obstacles. A set of obstacles is denoted as $P = P_1, P_2, ..., P_N$, and the center coordinate of the obstacles are denoted as $(x_1, y_1)$, $(x_2, y_2)$,...., and $(x_N, y_N)$ respectively. $d$ represents the minimum safe distance required between a UAV and an obstacle.

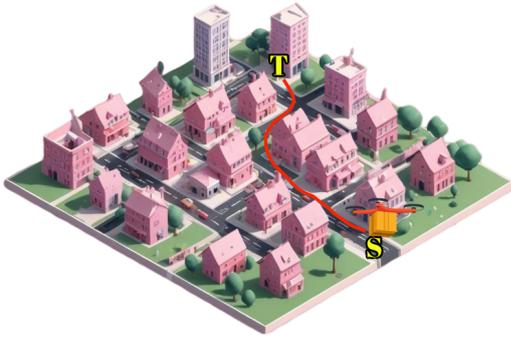

**Fig. 1.** UAV Delivery Path Example

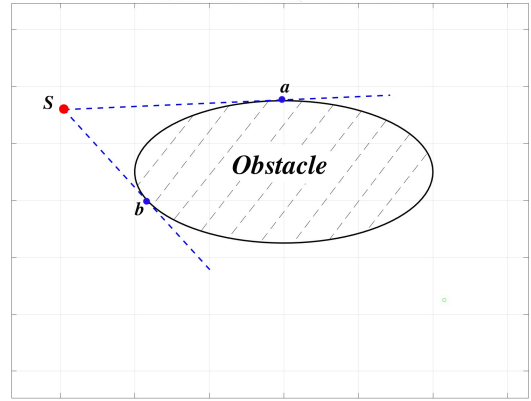

**Fig. 2.** Tangent Lines from Start Point S to Elliptic Obstacle

The UAV may start from $S$ to $T$, passing through several optimized node positions until it reaches the target point $T$ while avoiding obstacles present in the scene. The path-optimizing process must meet UAV constraints and requirements, including path length, total turning radius, algorithm execution time, and mission duration, which will be detailed later.

## 4. Design of Tangent Intersection Guidance (TIG)

Most traditional graph-based planning algorithms suffer from complexity and inefficiency because they need to construct paths for the entire map, especially when dealing with high-dimensional environments, leading to more computational resources. To solve this, the Tangent Intersection Guidance algorithm is introduced to generate multiple sub-paths towards the goal point. These sub-paths are derived based on various factors such as sub-path length, collided obstacles, smoothness, and other environmental considerations. However, rather than considering all generated sub-paths, the algorithm employs a heuristic function to select the most promising one, making it easier for the UAV to navigate with low computational resources.

In this paper, we employ two distinct algorithms: a static tangent planner (S-TIG) and a dynamic tangent planner (D-TIG). The environmental map is known in advance in the (S-TIG) planner, and obstacles remain static. This algorithm is suitable for pre-mission planning scenarios (Offline Planner). In contrast, the (D-TIG) planner is used in dynamic environments or situations where the map is partially known or completely unknown, such as scenarios involving exploration or rapidly changing environments where obstacle positions need real-time path planning execution (Online Planner).

This section briefly explains both static and dynamic planners. Since obstacles are represented as ellipses, each ellipse has two tangents from a given point $x$ outside the obstacle. For example, to execute a mission from a start point denoted as $S$, two tangent lines to the elliptic obstacle can be drawn at tangent points $a$ and $b$, respectively. Fig.2

Like the A* algorithm, the TIG Planner uses two sets: $CurrentSet$ and $ClosedSet$. The first set contains waypoints that are candidates for expansion, while the second set contains all explored waypoints. Each waypoint can result from an intersection of two tangent lines, which will be detailed in the next section. The algorithm procedure can be divided into two main functions. The first function is to get all candidate waypoints from the map. The second function is to choose the best subpath to integrate into the full path. All notations used in the algorithm are explained in Tab.1.

**Tab. 1.** Definitions of Main Notations

| Notations | Description |
|---|---|
| $S$ | Start-point |
| $T$ | Target-point |
| $W$ | A waypoint generated by the tangent planner |
| CurrentSet | A set to store candidate waypoints |
| ClosedSet | A set to store visited waypoints |
| treatedSet | A set that records tangent points that have already been calculated |

### 4.1. Static Tangent Planner (S-TIG) Algorithm Process

This algorithm generates a smooth, collision-free path for UAVs from S to T. Since the environmental map is already known, we first represent obstacles as ellipses using Eq.1. Each obstacle is defined by its position coordinates and major and minor semi-axes, which are used to calculate the tangent lines and generate candidate waypoints. The S-TIG planner procedure consists of two main steps: obtaining candidate waypoints and selecting the best waypoint for path generation. Initially, before the UAV mission begins, the start point is considered the current node in the first loop. Then, a straight line from this current node to the target point is calculated. However, this line is generally infeasible due to obstacles obstructing its way. Hence, two tangent lines are drawn and used as candidate sub-paths from the current node to the first encountered obstacle. The first obstacle encountered is the one closest to the current node intersecting with the direct line to the target node. This process is iteratively repeated until no obstacles are encountered using the tangent lines from the current node. The generation of waypoints can be defined as the intersection of two tangent lines at obstacle $O$; the first tangent line originates from the current node, and the second originates from the target point. However, this waypoint may prove insufficient due to multiple scenarios, potentially resulting in a path with poor smoothness or no feasible path. For example, as shown in Fig.3a, the waypoint $w$ is the intersection between the start point $S$ and the Target point $T$. Nonetheless, the subpath $SwT$ is unfeasible for UAVs to traverse. In the second scenario, as depicted in Fig.3b, the tangent line passing through $S$ is parallel to the tangent line passing through $T$, indicating no feasible path. In the last scenario, the tangent lines from $S$ and $T$ are impossible to intersect, meaning no waypoint can be added (Fig.3c).

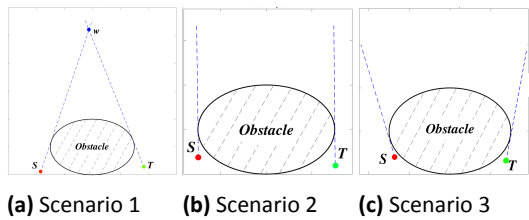

**(a)** Scenario 1　　**(b)** Scenario 2　　**(c)** Scenario 3

**Fig. 3.** 3 Potential Scenarios Resulting in Infeasible Paths

To address this, virtual ellipses around each obstacle are created using the following equations:

$$\begin{aligned}a_{\text{vir}} &= a + d \\ b_{\text{vir}} &= b + d\end{aligned} \quad (2)$$

Where $a$ and $b$ represent the semi-major and semi-minor axes of the obstacle respectively, and $a_{\text{vir}}$ and $b_{\text{vir}}$ represent the semi-major and semi-minor axes of the virtual ellipses. $d$ represents the safe distance between the obstacle and the virtual ellipse.

The waypoint is then determined as the intersection point between the virtual ellipse's perimeter and the obstacle's tangent line. It's important to note that there are two intersection points, and the selected waypoint is specifically chosen based on a criterion: it is the intersection point where the distance between the current point and the tangent point is smaller than the distance between the current point and the intersection point. Generally, this corresponds to the second intersection point from the current node along the subpath. As shown in Fig.4, two tangent lines are generated from the current node $N$. The first tangent line intersects with the red-dashed virtual ellipse surrounding the obstacle at two points. Our approach selects $w1$ as a new waypoint, which is the farthest intersection point from $N$. The same criterion applies to the second tangent line, choosing $w2$ as a second waypoint.

This approach ensures the generated path maintains smoothness and avoids unnecessary sharp turns.

After calculating and obtaining all current node candidate waypoints, a heuristic function is applied to select the best sub-path. Details of this heuristic function will be provided later.

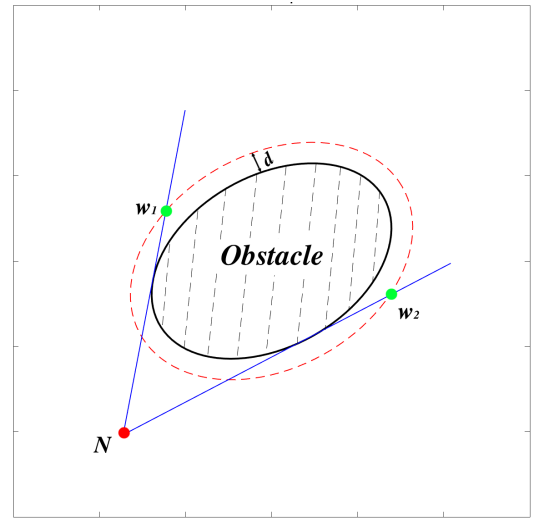

**Fig. 4.** Waypoint Generation Using Virtual Ellipse Technique

As illustrated in Fig.5a, the start point and the target point are denoted by $S$ and $T$ respectively, and $B_1, B_2, \ldots, B_11$ represent elliptic obstacles in the environment. We observe that the straight path from $S$ to $T$ is obstructed by obstacle $B_1$, which is the first collided obstacle. In this situation, the algorithm, as mentioned before, generates two tangent lines from $S$ to obstacle $B_1$ at points of tangency $P_1$ and $P'_1$, respectively. The algorithm also checks if both $SP_1$ and $SP'_1$ are clear subpaths, which is false in our example, so we handle each case separately. Firstly, we check the first collided obstacle with the tangent line $SP_1$ and generate two tangent lines, namely $SP_2$ and $SP'_2$, respectively. We can see now that both tangent lines $SP_2$ and $SP'_2$ are clear subpaths, so we create two waypoints $w_2$ and $w'_2$ using the virtual ellipse technique around obstacle $B_2$(Fig.5b). The same procedure is followed for the subpath $SP'_1$; drawing two tangent lines for obstacle $B_3$ to avoid it, we see that the subpath $SP_3$ is clear, so waypoint $W_3$ is created(Fig. 5c). For the last tangent line $SP'_3$, obstacle $B_4$ is in its way, so the tangent lines, namely $SP_4$ and $SP'_4$, are created, and waypoints $w_4$ and $w'_4$ are generated (Fig.5d). After extracting all candidate waypoints, the algorithm uses the heuristic rule to determine the best for the first subpath. After this, the current point is transferred as the selected waypoint in our example, which is $w_2$. The other subpaths are iteratively created using the same method until the current node reaches the end node. Finally, the algorithm extracts the final path from the closed set using each node with its previous node. The resultant path in this example is $S \to W_2 \to W_1 \to W_5 \to T$ (Fig.5e).

To choose the best waypoints, we apply the following heuristic rule:

$$H(w) = D(N, w) + \alpha P + D(w, T) \quad (3)$$

where $P$ represents the number of obstacles intersecting the tangent line from the current point $N$ to the waypoint $w$, $D(N,w)$ is the path length from $N$ to $w$, and $D(w,T)$ is the distance from $w$ to the target point $T$.

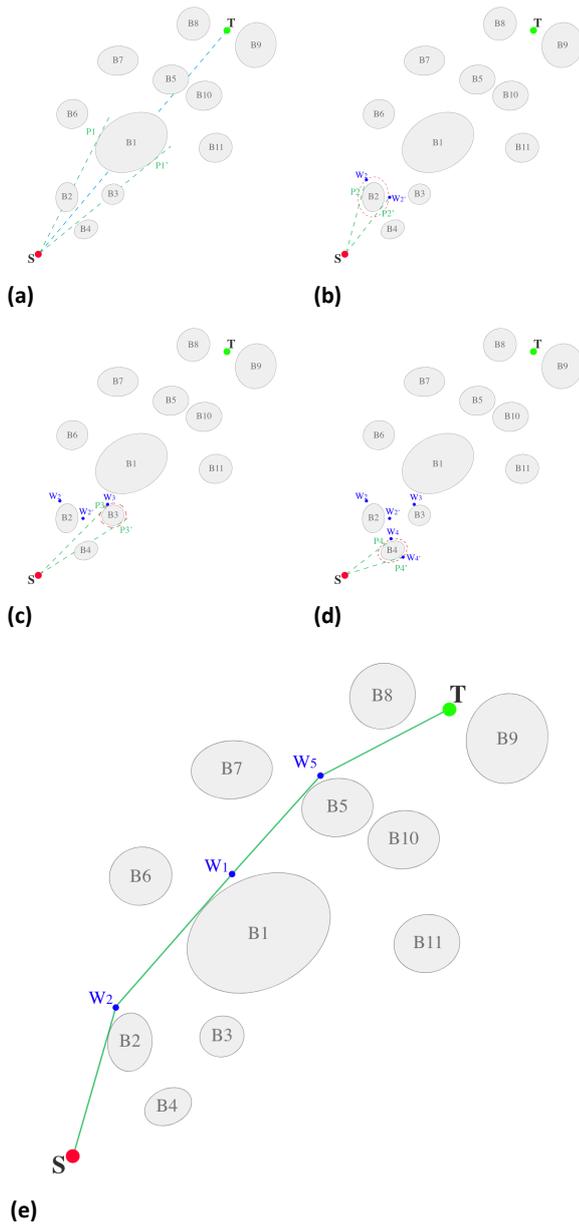

**Fig. 5.** S-TIG Algorithm Steps

Since our algorithm prioritizes a direct path to the goal, incorporating the obstacle count $P$ into the heuristic function encourages waypoint selection with fewer obstacles in way. This approach significantly reduces the number of obstacle interactions, leading to shorter, smoother, and computationally efficient trajectories. In addition, to balance the influence of $P$ while maintaining heuristic admissibility, we introduce the parameter $\alpha$, which adjusts its weight in the cost function. This refinement ensures that the algorithm remains both optimal and feasible, particularly in complex environments.

The steps of the S-TIG algorithm are detailed as follows:

1) Obstacle Modelization: Model the obstacles in the environment using Eq.1.

2) Initialization: Add the start point $S$ to the currentSet set and initialize the current node as the node with the minimum heuristic value in the currentSet, which is $S$.

3) Node Initialization: Initialise the current node in the $to\_Explore$ set.

4) Exploration Loop: While the $to\_Explore$ set is not empty, the algorithm selects the first node to be explored as a temporary target node $T_{temp}$.

5) Subpath Check: Determine whether the direct line from the current node to the temporary target point $T_{temp}$ is clear. If clear, generate a waypoint using the waypoint creation technique and append it to the $waypoints$ set. Otherwise, generate two tangent lines from the current node to the first collided obstacle, and add the points of tangency to $to\_Explore$ set.

6) Node Handling: Remove the temporary target node from $to\_Explore$ and add it to the *Explored* set.

7) Subpath Resolution: Relaunch Step. 4) if there are still no clear subpaths.

8) Waypoint Addition: Calculate heuristic values for each waypoint using Eq.3 and add them to *currentSet*.

9) Target Check: Check if $N$ it reaches the target point $T$. If it does, stop the algorithm and calculate the path; otherwise, return to Step. 3) for further exploration.

The algorithm pseudocode is shown in Alg.1:

### 4.2. Dynamic Tangent Planner (D-TIG) Algorithm Process

Unlike the S-TIG algorithm, the D-TIG is suitable for two situations: one in a partially known environment with unexpected obstacles and the other in a completely unknown environment.

**Dynamic Tangent Planner in a partially known environment**  In such scenarios, the UAV module initially employs the static planner to calculate the path. It then relies on UAV sensors to detect changes in the environment or obstacles' positions. If an obstacle appears or changes its coordinates, the static path becomes infeasible, and the dynamic planner recalculates the subpath affected by this obstacle. This ensures that the initial path remains unchanged except for the collided subpaths. As shown in Fig.6, the static offline planner generates an initial feasible path from point S to T, which is $S \rightarrow W_2 \rightarrow W_1 \rightarrow W_5 \rightarrow T$ in our example. The UAV begins its mission and travels from point S along the path toward the target point T. Integrated sensors continuously gather real-time information about the environment and detect changes. In the figure, an unexpected obstacle (B12) is detected by the UAV, making the subpath $W_5 \rightarrow T$ collide with B12. This obstacle renders the initial static path infeasible. Here, the dynamic planner's role comes into play: it replans only the collided subpath remains $W_5 \rightarrow T$ using the same static planner technique, providing a

**Algorithm 1** S-TIG Planner Algorithm
**Input:**
Start Node **S**, Target Node **T**
**Output:**
Path **Path**

1: Generate external ellipse for each obstacle using Eq.1
2: $currentSet \leftarrow \{S\}, closedSet \leftarrow \emptyset, treatedSet \leftarrow \emptyset$
3: **while** $currentSet$ is not empty **do**
4:    Get the node with the minimum heuristic value from $currentSet$ as the current node $N$.
5:    Add $N$ to $closedSet$
6:    **if** current node $N$ not equal to $T$ **then**
7:       Initialise $Explored \leftarrow \emptyset$, $to\_Explore \leftarrow \{T\}$ and $Waypoints \leftarrow \emptyset$
8:       **while** $to\_Explore$ is not empty **do**
9:          Let the temporary target node $T_{temp}$ be the first element in $to\_Explore$
10:          **if** the line segment from $N$ to $T_{temp}$ is a clear path and the angle is less than $\alpha$, and $T_{temp}$ is not in $treatedSet$ **then**
11:             Set $N$ as the parent of $T_{temp}$
12:             Add $T_{temp}$ to $Waypoints$
13:          **else**
14:             Generate two tangent lines of the first collided obstacle from the current node $N$
15:             Calculate each waypoint using virtual ellipse strategy using Eq.2 and add them to $to\_Explore$
16:             Delete $T_{temp}$ from $to\_Explore$ and add it to $Explored$
17:          **end if**
18:       **end while**
19:       Add all $Waypoints$ nodes to $currentSet$ with their heuristic values using Eq.3.
20:    **else**
21:       Calculate the path from $T$ to $S$ using nodes in $closedSet$ with their parents.
22:       Reverse the path to obtain the path from $S$ to $T$.
23:       **return** $Path$
24:    **end if**
25: **end while**

new feasible subpath $W_5 \rightarrow W_{12'} \rightarrow T$ to replace the infeasible one. This operation continues until the UAV safely reaches the target point T. The main steps of the

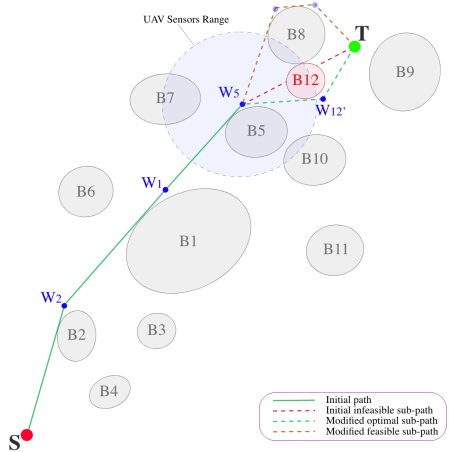

**Fig. 6.** Generated Path Using Dynamic Path Planner In Partially-known Environment

dynamic tangent planner in a partially known environment are as follows:

1) Running S-TIG: Use the Start point $S$ and the target point $T$ to Run the static tangent planner in Alg.1
2) Iterate the static planner path nodes
3) UAV Moving: Move UAV to the next node
4) Environment Checking: Check if there are changes in the environment with the current range. If so, the current node $N$ becomes the new start Node $N$ and reruns the static planner; otherwise, continue to the next node in the path.
5) Target Check: Stop the algorithm if the target point is reached; otherwise, go to Step.2)

D-TIG planner algorithm in partially-known environment pseudocode is shown in Alg.2:

**Dynamic Tangent Planner in completely unknown environment**   In the second situation, the environment is completely unknown except for the starting and target points. In this scenario, the algorithm plans the path only for the sub-environment detected by UAV sensors, using the S-TIG planner in each update while considering environmental changes. There are two cases when the environment is unknown: firstly, when sensors capture some obstacles within their limited range, and secondly, when no obstacles are in range. As shown in Fig.7a, only the starting point $S$ and the target point $T$ are known. Initially, the UAV captures obstacle positions in the sub-environment depending on the sensor's range (limit distance). The algorithm employs the static planner to avoid the first collided Obstacle $B1$ by creating tangent

**Algorithm 2** D-TIG Planner Algorithm in Partially Known Environment

**Input:**
Start Node **S**, Target Node **T**, Sensor Range radius **R**
**Output:**
Path **Path**

1: Run Static Path Planner Using Alg.1 and return the initial path set $P$ without the start point $S$
2: **while** true **do**
3:    $N \leftarrow$ the first element from $P$.
4:    Delete $N$ from $P$
5:    **if** $N$ is out of Range $R$ **then**
6:        $I \leftarrow$ the intersection between the sensor range perimeter $R$ and the line segment from the UAV's current position and $N$
7:        $N \leftarrow I$
8:    **end if**
9:    Move UAV to $N$
10:   **if** $N$ reaches the target point $T$ **then**
11:       **Break the loop**
12:   **end if**
13:   Update Environment information
14:   **if** the environment has changed **then**
15:       Run Static Path Planner Using Alg.1 from $N$ as a new start node and return the updated path $P$ without $N$
16:   **end if**
17: **end while**

---

lines and employs the heuristic function to generate a new waypoint $W1$. While the target point is not yet reached, the algorithm generates a direct line to $T$. The problem is that the environmental information is missing because $T$ is out of range. In this case, the algorithm creates a maximum range waypoint within the range perimeter. These waypoints are used when the next node is out of range. The maximum range waypoint is defined as the intersection between the sensor perimeter and the direct line from the previous waypoint $W1$ to $T$. Additionally, the UAV moves to the latest maximum range waypoint, which is $W2$ as shown in Fig.7b. In this case, the sensors detect no intersecting obstacles between $W2$ and the range perimeter. Consequently, the D-TIG planner creates another maximum range waypoint $W3$ and makes the UAV move to it, gathering environmental information along the way. As depicted in Figs. 7c and 7d, the D-TIG planner continues avoiding obstacles and creating maximum range waypoints if necessary until the target point is reached. The final planned path is in Fig.7e.

The main steps of the dynamic tangent planner for completely unknown are as follows:

1) Obstacele Modeling: Use Eq.1 to model the obstacles in the sub-environment.

2) Initialization: Begin by initializing sets and variables, including *currentSet* for unexplored nodes, *treatedSet* for calculated waypoints, and $T_{temp}$ as a temporary target node.

3) Exploration Loop Initialization: Begin a loop to explore until the *currentSet* is empty.

4) Select Current Node: Select the node with the minimum heuristic value from the *currentSet* as the current node $N$.

5) Check Range: If the current node $N$ is exactly in the range perimeter $R$, update the path, move UAV, and update environment information. Then reinitialize *currentSet* with $N$.

6) Continue Exploration: If the *currentSet* is not empty, continue exploration.

7) Check Current Node: If the current node $N$ is not reached the target node $T$, proceed with waypoint exploration.

8) Waypoint Exploration: Explore waypoints from the current node $N$ to the target node $T$.

9) Check Clear Subpath: Check if the line segment to the temporary current node $N_{temp}$ is clear and if it has not been treated before.

10) Add Waypoints: Calculate heuristic values for each waypoint using Eq.3 and add them to *currentSet*.

11) Check Target Node: If the current node $N$ is the target node $T$, terminate the algorithm and extract the final subpath.

D-TIG planner algorithm in unknown environment pseudocode is shown in Alg.3:

### 4.3. Path Smoothing Technique

Generated paths using static or dynamic planners often include a series of connected waypoints, but due to their complexity and flight dynamic requirements, they may not be suitable for UAVs. Several research papers have proposed techniques such as Bezier, spline, and Dubin's Curves to smooth these paths and address this issue [41]. However, these methods are effective only in open spaces or situations where the path waypoints are not too close to obstacles presented

**Algorithm 3** D-TIG Planner Algorithm In Unknown Environment
**Input:**
Start Node **S**, Target Node **T**, Sensor Range radius **R**
**Output:**
Path **Path**

1: $N \leftarrow S$
2: **while** $T$ is out Of Range $R$ **do**
3:     Initialize *currentSet* $\leftarrow \{N\}$, *treatedSet* $\leftarrow \emptyset, T_{temp} \leftarrow T$
4:     **while** $currentSet$ is not empty **do**
5:         Get the node with the minimum heuristic value from *currentSet* as the current node $N$
6:         Delete $N$ from *currentSet*
7:         **if** $N$ is exactly in the range perimeter $R$ **then**
8:             Calculate in range subpath starting from $N$
9:             Move UAV following the obtained subpath
10:             Update Environment Information
11:         **end if**
12:         Initialize $Explored \leftarrow \emptyset, to\_Explore \leftarrow \{N\}, Waypoints \leftarrow \emptyset$
13:         **while** $to\_Explore$ is not empty **do**
14:             Let temporary target node $T_{temp}$ be the first element in $to\_Explore$
15:             Delete $T_{temp}$ from $to\_Explore$ and add it to $Explored$
16:             **if** the line segment from $N$ to $T_{temp}$ is a clear path and the angle is less than $\alpha$, and $T_{temp}$ is not in $treatedSet$ **then**
17:                 **if** $T_{Temp}$ is out of range $R$ **then**
18:                     Move $T_{Temp}$ in range using the intersection between $R$ perimeter and the line $NT_{Temp}$
19:                 **end if**
20:                 Set $N$ as the parent of $T_{temp}$
21:                 Add $T_{temp}$ to $Waypoints$
22:             **else**
23:                 Generate two tangent lines of the first collided obstacle from $N$.
24:                 Calculate each waypoint using virtual ellipse strategy using Eq.2 and add them to $to\_Explore$
25:                 Delete $T_{temp}$ from $to\_Explore$ and add it to $Explored$
26:             **end if**
27:         **end while**
28:         Add all waypoints to *currentSet* with their heuristic values using Eq.3
29:     **end while**
30: **end while**

in the environment, meaning there is a high risk of collision in cluttered and dense environments. As depicted in Fig. 8, the initial path is not yet smoothed. A new smoothed path is obtained by applying quadratic Bezier curves to the path, which unfortunately still collides with some obstacles. Fig.9 shows examples of collision areas highlighted within red dashed circles.

To solve this, the paper proposes using a quadratic Bezier curve technique with every three successive nodes from the start to the target node. Each waypoint, except the start and target nodes, is considered a turn. The idea is to create a curve that smooths these turns, making it suitable for UAVs. Firstly, we extract each waypoint on the path and create two temporary waypoints before and after each waypoint and handle them as the first and last control points for the quadratic Bézier curve, respectively. Secondly, the algorithm uses the quadratic Bezier curve to smooth these waypoints and combines the collected curves in a final smoothed path. As shown in Fig. 10a, the first temporary waypoint $A'$, the waypoint A, and the second temporary point $A''$ are utilized as control points.

After applying the quadratic bezier curve, a smooth subpath is obtained (Fig. 10b). This process continues until the target node is reached.

## 5. Experimental Setup

This section evaluates the path planning algorithm's performance and effectiveness in static and dynamic environments. The paper conducts multiple simulation experiments to compare it with other popular path planning algorithms, such as A*, PRM, RRT*, Tangent Graph, and APPATT(SETG-TG) in static environments and APF and APPATT(DETG-TG) [42] in dynamic environments. These algorithms are coded using MATLAB R2021b on a MacBook Pro 2020 with an Intel i5 processor and 8GB of memory. During the experiment, the path planning process must comply with specific constraints, such as minimizing the path length, execution time, and enforcing a minimum turning radius, to guarantee a low number of curves and an efficient trajectory. As mentioned in Section 2, this study did not consider the height of obstacles presented in the environment and assumed the map to be two-dimensional. Additionally, to minimize the com-putation time of the Tangent Graph algorithm, we calculate the waypoints of the algorithm based on the intersection points of the tangent lines from the current and end positions. Moreover, we assume the UAV range distance is 60m.

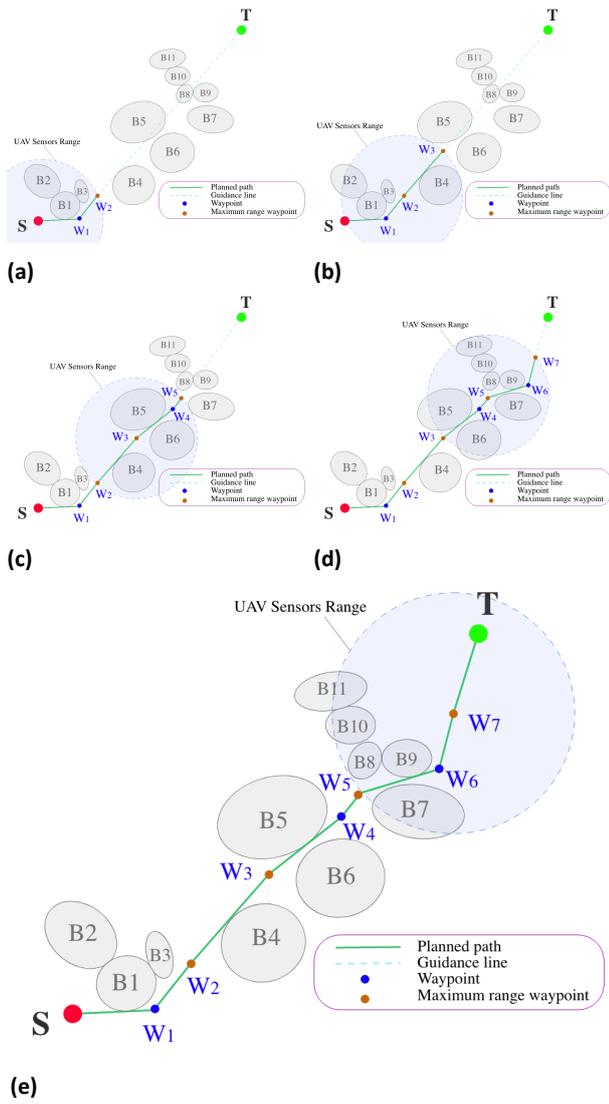

**Fig. 7.** D-TIG Planner Algorithm Steps In Unknown Environment

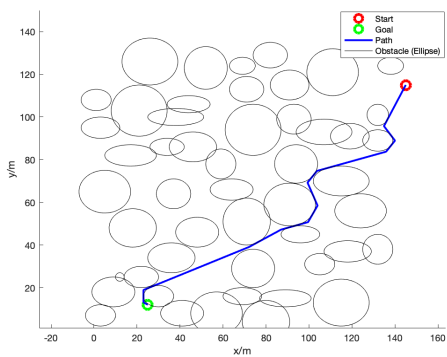

**Fig. 8.** Example Of a Generated Path Without Smoothing

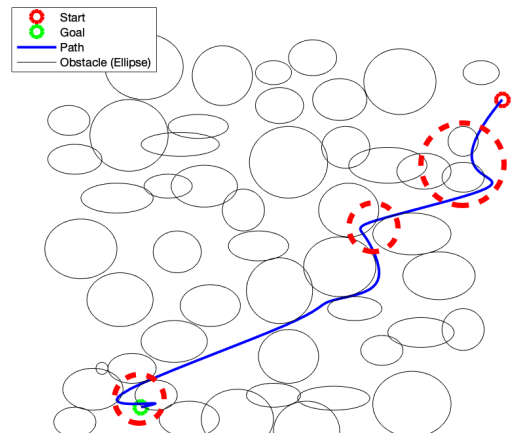

**Fig. 9.** A Smoothed Path Using Quadratic Bézier Curve With Collision

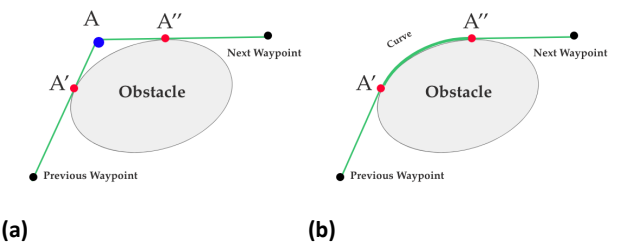

**Fig. 10.** TIG Smoothing Steps

### 5.1. Environmental Modeling

UAVs may encounter multiple scenarios in real-world operations. For example, they may need to avoid a single obstacle in some cases, while in others, they must navigate through a densely obstructed area. Similarly, they may have long travel distances in some missions and short travel distances in others. To address these variations, this paper examines four experiments, as follows:

**Short Map vs. Large Map**    In these two experiments, we evaluate whether the algorithm can generate feasible paths in environments of different sizes. The short map in this paper measures 500 × 500 m, while the large map is 1000 × 1000 m. These experiments ensure that the algorithms can balance between path and turning angles optimality and computational efficiency.

**Sparse vs. Dense Maps**    The number of obstacles in a path planning process plays a crucial role. The first experiment involves planning a path in a densely obstructed area, where finding a feasible route for the UAV is significantly challenging. In contrast, the sparse map presents a less demanding environment with fewer obstacles. These experiments assess the algorithm's performance across different urban conditions, highlighting its adaptability and efficiency in navigating environments of varying spatial complexity. In this paper, we define a sparse area as one con-taining 10% of obstacles, while a dense area contains more than 60% of obstacles.

### 5.2. Evaluation Metrics

After a literature review of many studies, the proposed algorithms employ only path length minimization as an objective function to maximize the operational range of UAVs. However, this approach can result in paths with sharp turns and unnecessary nodes, negatively impacting energy consumption and path efficiency. Hence, this paper adopts objective functions to minimize the path length, turning radius, and execution time. This approach aims to generate smoother, more adaptable paths that optimize mission efficiency while ensuring operational feasibility.

**Short Path Objective Function**   Minimizing flight distance is crucial for a UAV path planning algorithm since it directly reduces travel time and energy consumption. Therefore, an objective function is defined to prioritize the shortest path length as follows:

$$min D(x,y) = \sum_{i=1}^{n-1} \sqrt{(x_{i+1}-x_i)^2 + (y_{i+1}-y_i)^2} \quad (4)$$

where $n$ the length of the path, $(x_i, y_i)$ and $(x_{i+1}, y_{i+1})$ are the abscissa and ordinates of the $i$-th node, and $(x_{i+1}, y_{i+1})$ $(i+1)$-th node respectively. The Euclidean distance is used to calculate the path's length.

**Minimum Total Turning Angles Objective Function** Each turning angle impacts the overall smoothness of the path. The lower the sum of turning angles, the smoother the path we get. In Contrast, a higher sum results in high energy consumption and requires more motion time. Therefore, an objective function to minimize the turning angles is adopted as follows:

$$\min T_r(\theta) = \sum_{i=1}^{n-2} \theta_i. \quad (5)$$

$$\theta_n = \left| \arctan \left( \frac{\frac{y_{n+1}-y_n}{x_{n+1}-x_n} - \frac{y_{n+2}-y_{n+1}}{x_{n+2}-x_{n+1}}}{1 + \frac{y_{n+1}-y_n}{x_{n+1}-x_n} \cdot \frac{y_{n+2}-y_{n+1}}{x_{n+2}-x_{n+1}}} \right) \right|. \quad (6)$$

Where $\theta_n$ is the angle at the $n$th node, which is calculated based on its previous node $n-1$, and its next node $n+1$.

**Shortest Algorithm Execution Time**   The execution time of the algorithm also has a significant impact, especially in dynamic environments where UAVs need to adjust their flight paths to avoid obstacles or achieve mission objectives. Adopting a function to minimize algorithm execution time ensures that decisions are made swiftly, contributing to faster and more responsive navigation. The objective function is set as follows:

$$\min T(a) = t \quad (7)$$

Where $t$ denotes the algorithm $a$ execution time.

### 5.3. Simulation Experiment

**Static Path Planner Simulation Experiment**   Based on the simulation results shown in Figs. 11–14 and the summarized performance metrics in Tab. 2, it's evident that the S-TIG algorithm consistently outperforms A*, PRM, RRT*, Tangent Graph (TG), and AP-PATT algorithms across all experiments with randomly generated obstacles of varying numbers and sizes. Firstly, the static planner reduces path length by 5% compared to the A*, 6.98% compared to PRM and a substantial 22.80% compared to RRT*, 7.55% compared to APPATT algorithms , and approximately the same with the tangent graph planner with a difference of 0.18%. Additionally, the execution time of the S-TIG planner algorithm averages a reduction of 98.08%, 98.29%, 47% , 7% and 88.25% compared to A*, PRM, RRT*, Appatt, and Tangent Graph, respectively. In terms of the sum of turning angles, S-TIG demonstrates a reduction of 93.39% compared to A*, 65.15% compared to PRM, 84.27% compared to RRT*, 35.41% compared to APPATT while S-TIG produces the same turning angles as the tangent graph planner in most cases. These results indicate that the S-TIG algorithm consistently generates shorter paths with fewer nodes, requiring less time and producing smoother trajectories due to the minimized total turning angles. From the APPATT (SETG-TG) test results, the algorithm failed due to its structure and the type of environment, especially in C2, C7, and C12, while it also failed in all cases in dense environments.

The APPATT algorithm employs an intersection strategy, which leads to many failures, as represented in Fig. 3. In contrast, S-TIG addresses this limitation by adopting the waypoint creation technique. Additionally, APPATT does not implement a strategy for checking created waypoints when they become stuck in overlapping obstacles, which can result in an infinite loop when attempting to choose the best waypoint. Due to these limitations, APPATT failed the tests across five dense maps, whereas the TIG algorithm demonstrated strong capabilities in path generation within static environments.

The A* algorithm can produce an optimal path based on grids with 1m x 1m resolution. However, the quality of the generated path is not satisfactory. In addition, A* and PRM algorithms suffer from high computation times, especially in large areas. For example, in C6, A* needs 98 seconds to generate a feasible path, while PRM takes 5 seconds. RRT*, due to its randomness, generates a large number of turning angles, sometimes exceeding 40 radians. This algorithm cannot guarantee path optimality and smoothness for UAVs. This often results in UAVs having to take sharp turns or unnecessary nodes, which can impact the overall mission time. The tangent graph planner produces the shortest paths with the minimum sum of turning angles; however, this algorithm calculates the tangents for the entire map, leading to higher computation times, especially in dense areas.

In contrast, the S-TIG algorithm always evaluates the subpath quality using its heuristic rules be-fore making a decision, which eliminates any sharp turns and redundant nodes and produces collision-free paths with shorter times in each of the cases in all scenarios. Next in line is Tangent Graph, followed by Appatt, A*, PRM, and RRT*. To sum up, the S-TIG static planner algorithm shows a superior path planning performance compared to the other five algorithms in terms of path length, time consumption, and turning angles in static environments.

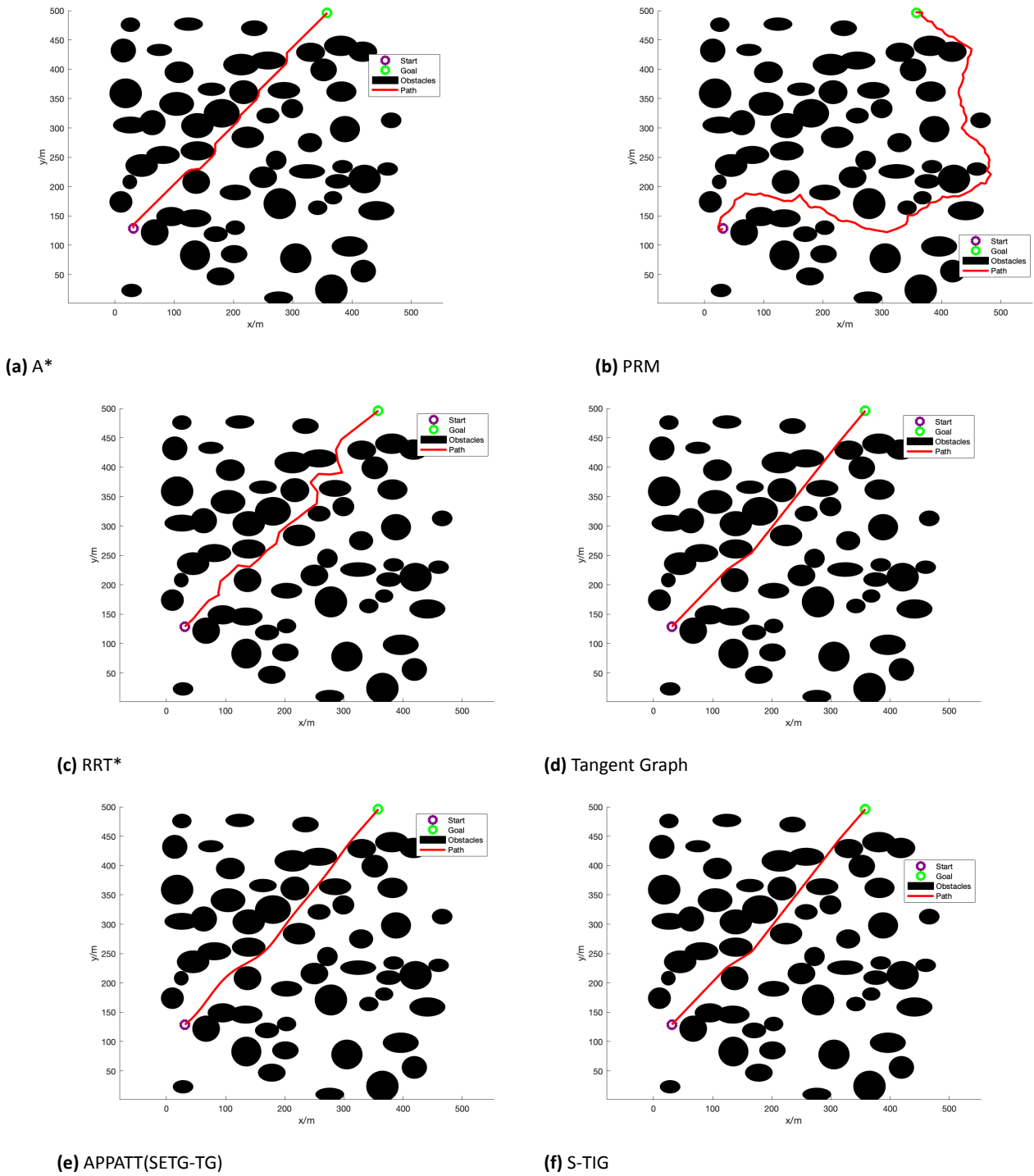

**Fig. 11.** Generated paths in static environments on a short map (C1)

**Dynamic Planner Simulation Experiment In Unkonwn Environment** To test the effectiveness of the D-TIG planner in an unknown environment, this paper compares the algorithm with two dynamic planners: APF and APPATT. Figures 15 to 18 show the planned paths for four scenarios with varying obstacle sizes and positions. The test results, summarized in Tab. 3, show that D-TIG consistently reduced the planned path length by 20.55% compared to the APF algorithm and approximately 0.1% compared to APPATT in all cases.

Additionally, the algorithm's execution time is equal to that of APPATT, requiring only 0.01 seconds on average to plan a near-optimal route, unlike APF, which demands significantly more time, sometimes up to 0.14 seconds in high-dimensional environments (e.g., C21). Moreover, the D-TIG dynamic planner produces fewer turning angles than the other algorithms, reducing them by approximately 87% compared to APF and 34.11% compared to APPATT.

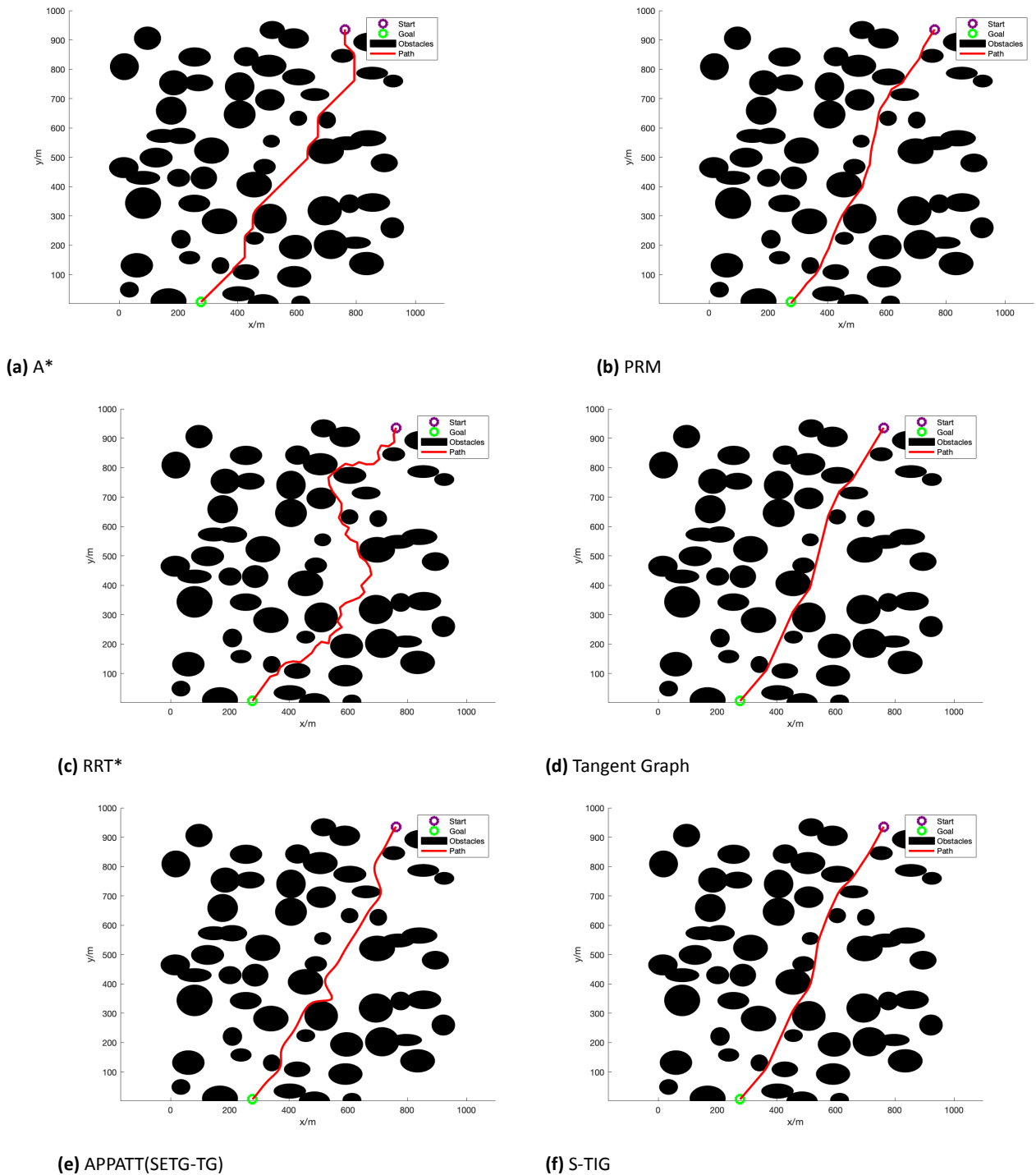

**Fig. 12.** Generated paths in static environments on a large map (C5)

Across all test scenarios, the D-TIG algorithm successfully generated paths, whereas the other algorithms failed in multiple cases, such as C20, C24, C29, C30, C31, and C32. In summary, the D-TIG planner is effective in unknown environments, generating shorter paths in less time and with fewer unnecessary turns than the other algorithms.

**Dynamic Planner Simulation Experiment In Partially Known Environment With Unexpected Obstacles** Another simulation experiments were conducted to test the effectiveness of the D-TIG algorithm in environments with unexpected obstacles. Figures 19 to 21 illustrate the initially planned path in red before the environmental change, and the green color represents the corrected path after the UAV faced unexpected obstacles, colored in orange. The D-TIG was compared with the APPATT algorithm since it is designed to calculate new paths based on unexpected obstacles, and the results were collected in Tab.4. The re-sults demonstrate that the D-TIG planner can produce shorter, smoother paths in less time. Additionally, the algorithm reduces the planned path length by 5.63% across all test cases and decreases the path turning radius by 25.55% compared to the APPATT algorithm. Moreover, the execution time of both algorithms is approximately the same, which both not exceed 0.08 seconds for replanning, which is good to allow UAVs to make decisions more quickly. Furthermore, the APPATT algorithm fails to generate feasible paths in multiple cases (e.g C21 and C26), which UAVs cannot rely on it in such situations because the risk of collision is too high in real-time scenarios. The APPATT algorithm also fails to find feasible paths in unknown environments, as shown in Tab. 4. Similarly, it cannot find a path in a dense environment with pop-up obstacles. In contrast, the D-TIG algorithm successfully handles dense maps with pop-up obstacles, as shown in Fig. 22. These improvements can be attributed to the fundamental differences between the two algorithms. The APPATT algorithm is based on the tangent intersection with the goal position to create feasible waypoints in the search space. In contrast, the dynamic planner (D-TIG) uses a waypoints generation technique, which produces paths closer to pop-up obstacles while maintaining a safe distance to avoid collisions. Based on these results, it is evident that the D-TIG is more effective in a partially known environment with unexpected obstacles compared to the APPATT algorithm.

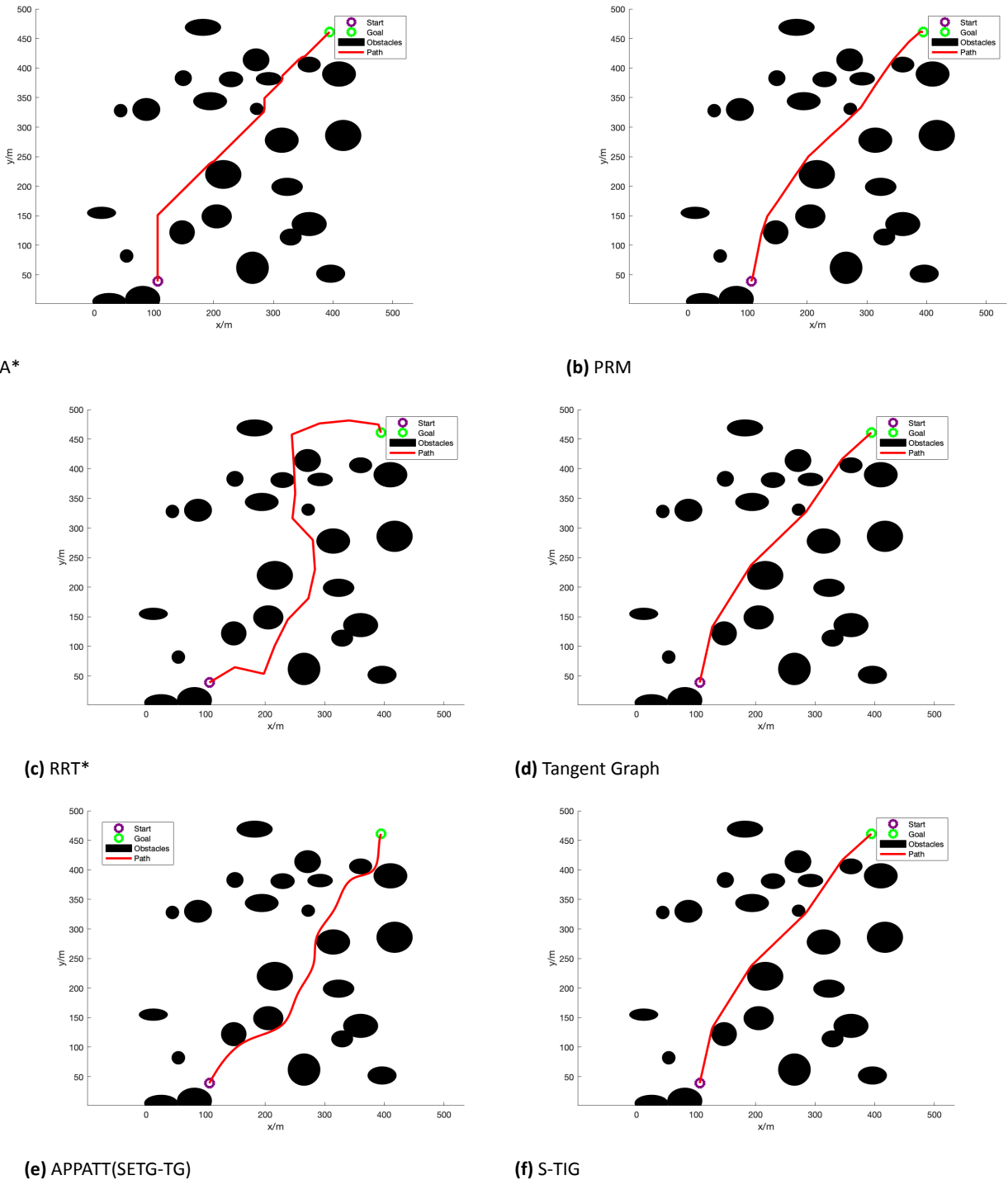

**Fig. 13.** Generated paths in static environments on a sparse map (C9)

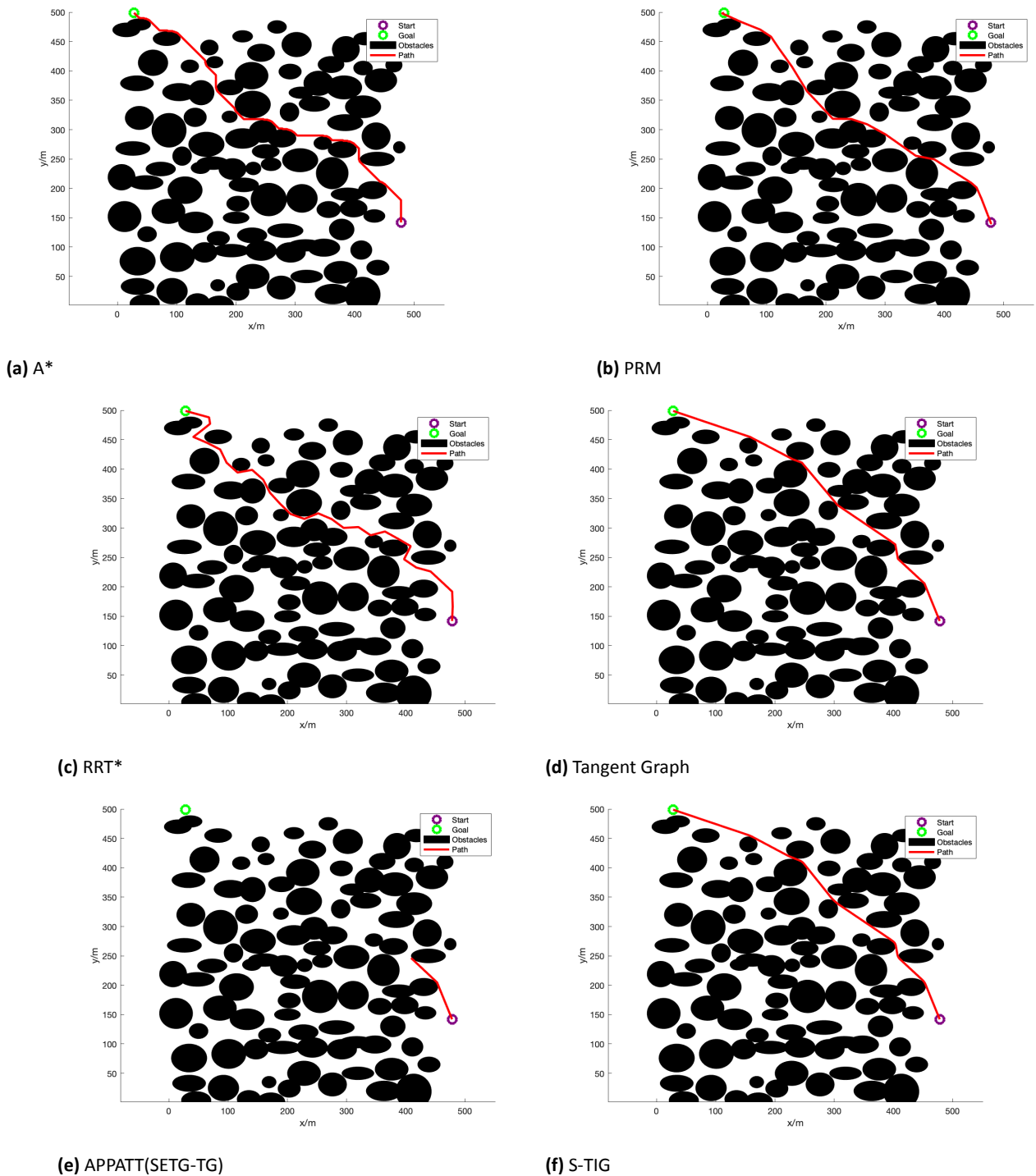

**Fig. 14.** Generated paths in static environments on a dense map (C13)

## 6. Conclusion

In this paper, we present the Tangent Intersection Guidance (TIG) algorithm, a novel approach for UAV path planning in both static and dynamic environments. The algorithm generates two sub-paths for each elliptic obstacle and selects the optimal one based on a heuristic rule. This process is iteratively repeated until the target point is reached. The static planner (S-TIG) is employed in known environments, and our test results demonstrate that S-TIG generates paths that are, on average, 11% shorter compared to other static methods while reducing the number of turns by 70% and maintaining a low computation time of around 0.1 seconds, even in higher-dimensional environments. Additionally, the dynamic planner (D-TIG) functions as a local planner in partially known environments with unexpected obstacles and completely unknown environments, outperform-

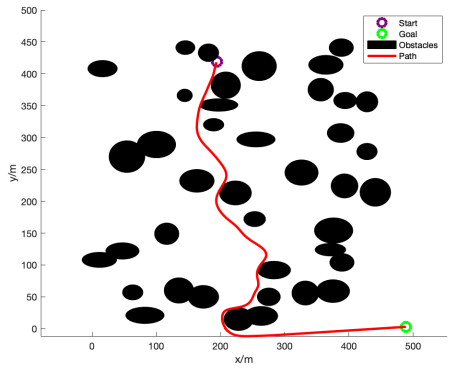

**(a)** APF

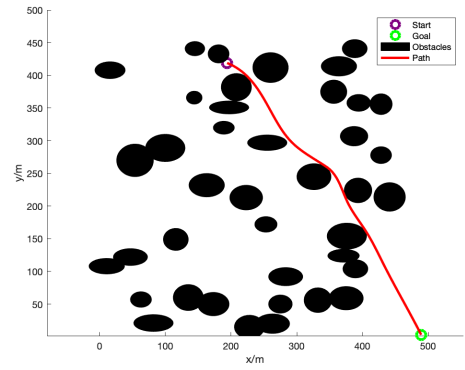

**(b)** APPATT(DETG-TG)

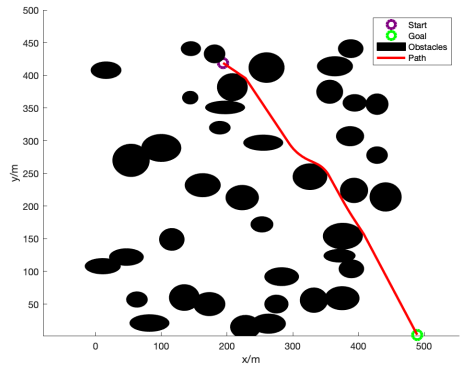

**(c)** D-TIG

**Fig. 15.** Generated paths in unknown environment on a short map (C17)

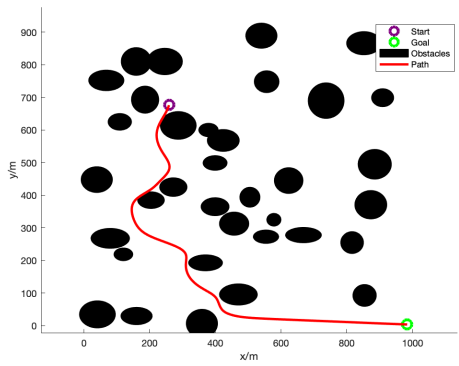

**(a)** APF

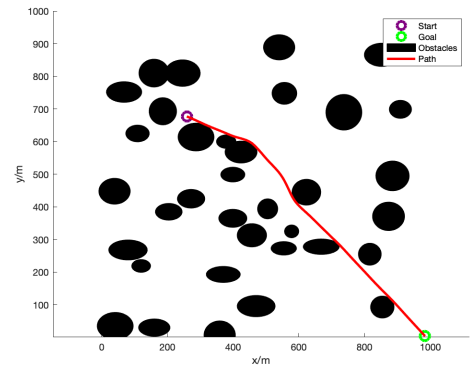

**(b)** APPATT(DETG-TG)

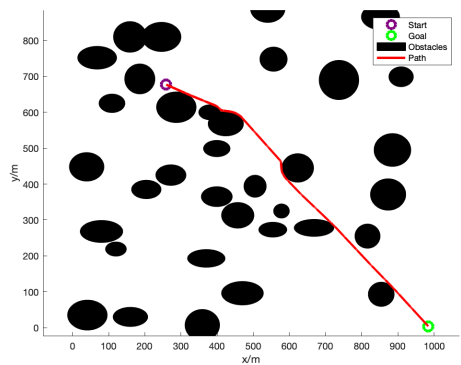

**(c)** D-TIG

**Fig. 16.** Generated paths in unknown environment on a large map (C21)

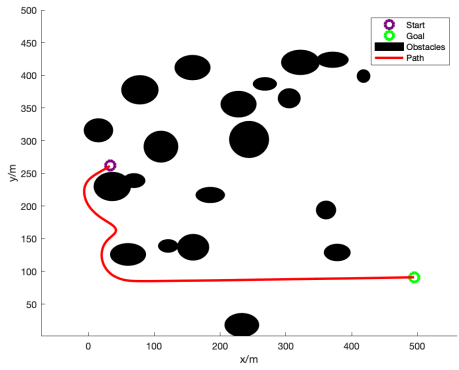

**(a)** APF

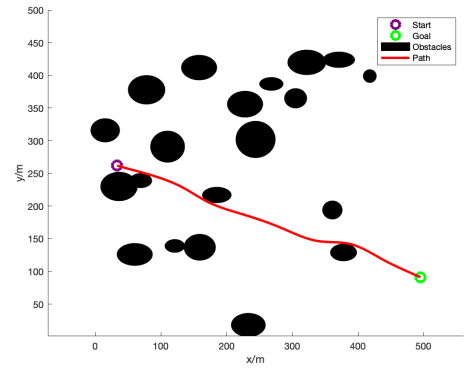

**(b)** APPATT(DETG-TG)

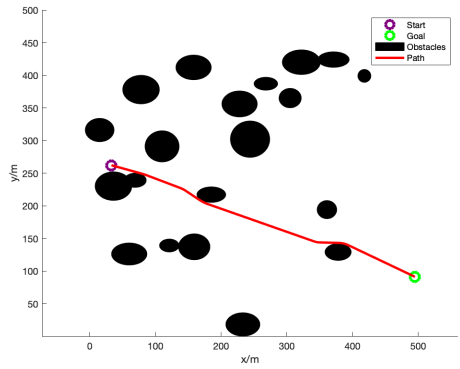

**(c)** D-TIG

**Fig. 17.** Generated paths in unknown environment on a sparse map (C25)

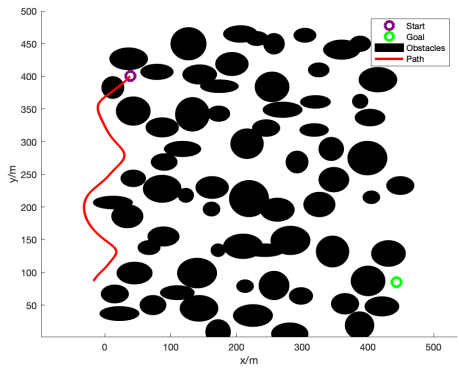

**(a)** APF

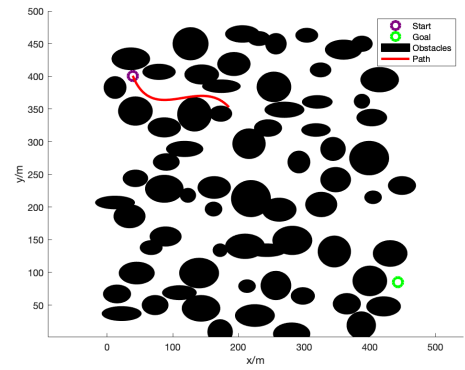

**(b)** APPATT(DETG-TG)

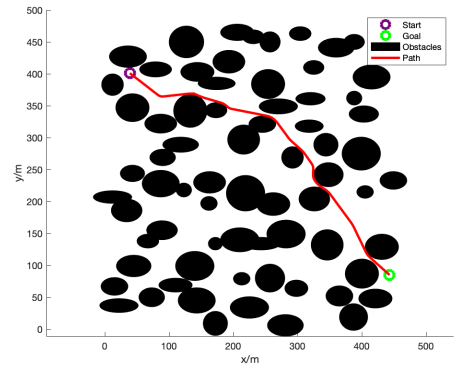

**(c)** D-TIG

**Fig. 18.** Generated paths in unknown environment on a dense map (C29)

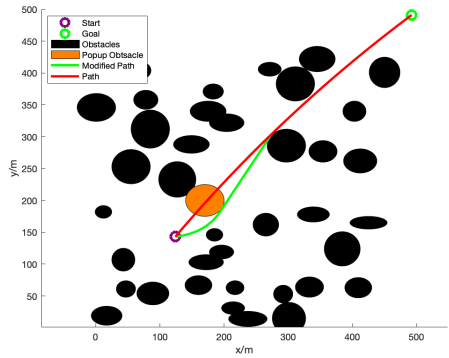
**(a)** APPATT

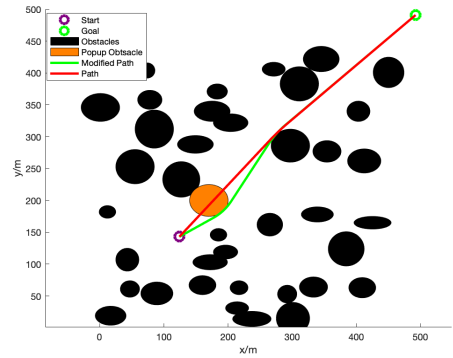
**(b)** D-TIG

**Fig. 19.** Generated paths in a partially known environment with pop-up obstacles on a short map (C19)

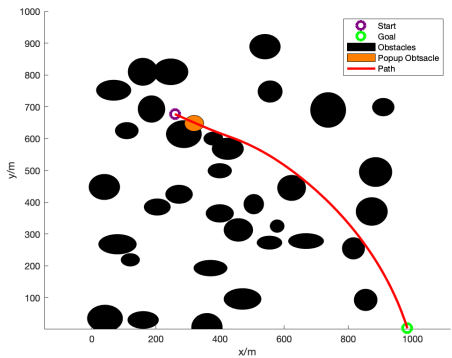
**(a)** APPATT

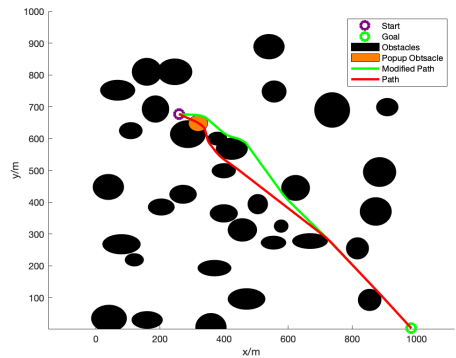
**(b)** D-TIG

**Fig. 20.** Generated paths in a partially known environment with pop-up obstacles on a long map (C20)

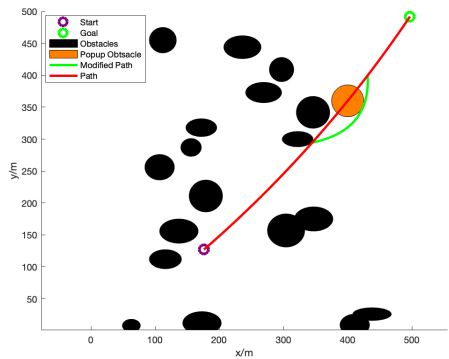
**(a)** APPATT

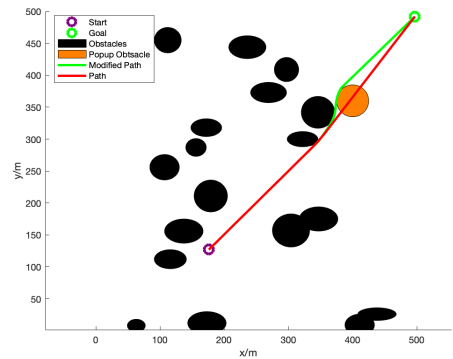
**(b)** D-TIG

**Fig. 21.** Generated paths in a partially known environment with pop-up obstacles on a sparse map (C25)

ing existing real-time algorithms by achieving fast replanning times under 0.08 seconds only, while reducing path length and turning angles by approximately 9% and 50%, respectively, compared to other dynamic algorithms. Despite its advantages, TIG has some limitations, particularly in guaranteeing path length optimality in dynamic environments. Moreover, the scalability of TIG to high-dimensional spaces, especially full 3D path planning, requires additional exploration. Future research will focus on extending the TIG framework to three-dimensional environments, optimizing heuristic cost functions to ensure admissibility, and refining path smoothing techniques for a fairer comparison with other algorithms. In conclusion, the Tangent Intersection Guidance algorithm represents a significant step forward in UAV path planning technology. By integrating heuristic-based decision-making and Bézier curve smoothing, TIG enhances UAV navigation efficiency and safety in complex environments. Future studies will focus on addressing the identified limitations, improving computational scalability, and benchmarking against recent advancements in visibil-

**Tab. 2.** Comparison of Different Static Path Planning Algorithms Across Four Scenarios

| Map Type | Case | Path Length | | | | | Time | | | | | Turning Radius | | | |
|---|---|---|---|---|---|---|---|---|---|---|---|---|---|---|---|
| | | S-TIG | A* | PRM | RRT | APPATT | TG | S-TIG | A* | PRM | RRT | APPATT | TG | S-TIG | A* | PRM | RRT | APPATT | TG |
| Short | C1 | 493.84 | 510.06 | 949.33 | 566.49 | 495.29 | 493.59 | 0.01 | 8.44 | 3.65 | 0.18 | 0.09 | 0.20 | 0.53 | 32.20 | 58.35 | 14.96 | 1.45 | 0.53 |
| | C2 | 511.89 | 533.16 | 661.20 | 595.68 | N/A | 511.60 | 0.07 | 11.72 | 3.65 | 0.27 | N/A | 2.73 | 0.68 | 32.20 | 42.18 | 16.84 | N/A | 0.46 |
| | C3 | 520.53 | 545.25 | 587.87 | 931.31 | 531.38 | 514.25 | 0.06 | 9.26 | 3.68 | 0.64 | 0.21 | 2.81 | 2.81 | 49.48 | 30.59 | 36.17 | 1.99 | 1.86 |
| | C4 | 527.59 | 557.87 | 645.85 | 840.31 | 584.34 | 528.50 | 0.06 | 10.38 | 3.36 | 0.14 | 0.06 | 2.02 | 2.34 | 46.33 | 37.94 | 26.21 | 2.72 | 2.34 |
| Large | C5 | 1063.13 | 1156.40 | 1071.65 | 1429.14 | 1135.71 | 1063.76 | 0.11 | 73.00 | 5.25 | 0.77 | 0.15 | 73.48 | 1.97 | 55.76 | 6.43 | 34.66 | 9.28 | 1.72 |
| | C6 | 1004.27 | 1022.01 | 1040.68 | 1313.74 | 1064.00 | 1004.14 | 0.02 | 92.65 | 5.09 | 0.35 | 0.10 | 0.14 | 0.16 | 10.21 | 7.20 | 34.58 | 1.87 | 0.14 |
| | C7 | 1241.35 | 1289.83 | 1253.50 | 1558.46 | N/A | 1238.23 | 0.05 | 80.09 | 5.88 | 0.78 | N/A | 4.13 | 1.49 | 69.90 | 6.69 | 44.56 | N/A | 1.07 |
| | C8 | 1001.01 | 1054.11 | 1019.10 | 1254.34 | 1925.46 | 999.66 | 0.09 | 72.85 | 5.25 | 0.39 | 0.14 | 8.57 | 1.67 | 52.62 | 5.08 | 35.32 | 10.29 | 1.42 |
| Sparse | C9 | 522.01 | 545.98 | 524.81 | 706.33 | 543.33 | 522.24 | 0.04 | 16.00 | 9.85 | 0.11 | 0.01 | 0.83 | 1.02 | 18.06 | 2.30 | 8.14 | 5.55 | 1.02 |
| | C10 | 515.36 | 548.42 | 526.48 | 714.74 | 515.30 | 515.38 | 0.03 | 17.47 | 9.02 | 0.02 | 0.03 | 0.04 | 0.25 | 14.92 | 4.76 | 12.66 | 0.24 | 0.24 |
| | C11 | 514.74 | 524.83 | 525.51 | 643.16 | 524.88 | 514.81 | 0.01 | 12.82 | 9.44 | 0.11 | 0.05 | 0.03 | 0.37 | 13.35 | 6.32 | 16.09 | 0.69 | 0.37 |
| | C12 | 511.61 | 520.28 | 525.26 | 613.53 | N/A | 509.00 | 0.06 | 16.69 | 9.36 | 0.09 | N/A | 0.16 | 2.05 | 16.49 | 6.08 | 15.28 | N/A | 1.14 |
| Dense | C13 | 601.13 | 652.94 | 613.57 | 713.55 | N/A | 602.23 | 0.32 | 6.71 | 10.90 | 0.34 | N/A | 57.50 | 2.97 | 69.90 | 9.38 | 17.87 | N/A | 2.96 |
| | C14 | 509.77 | 562.94 | 515.49 | 604.97 | N/A | 509.69 | 0.29 | 6.73 | 9.25 | 0.47 | N/A | 5.26 | 2.46 | 62.05 | 6.15 | 12.70 | N/A | 2.14 |
| | C15 | 686.06 | 698.61 | 638.75 | 868.91 | N/A | 686.05 | 0.68 | 6.16 | 11.60 | 0.42 | N/A | 62.59 | 18.90 | 68.33 | 13.63 | 23.39 | N/A | 18.90 |
| | C16 | 572.76 | 644.14 | 574.83 | 711.90 | N/A | 568.13 | 0.32 | 4.35 | 10.10 | 0.34 | N/A | 61.27 | 11.80 | 52.62 | 7.33 | 17.47 | N/A | 3.80 |

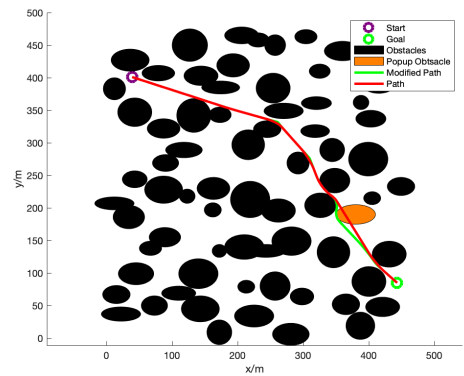

**(a)** D-TIG (C25)

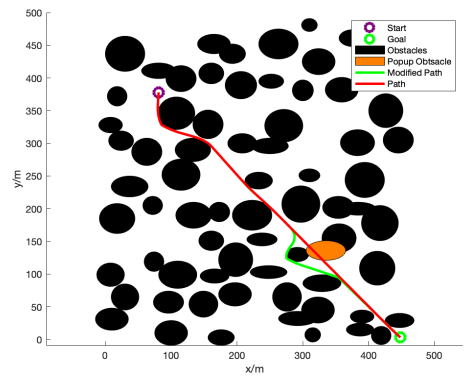

**(b)** D-TIG(C26)

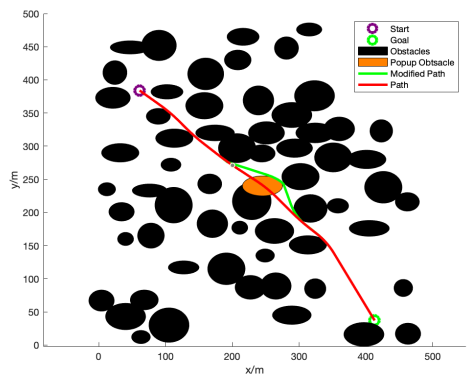

**(c)** D-TIG(C27)

**Fig. 22.** Generated paths using D-TIG in a partially known environment with pop-up obstacles on dense maps

ity graph-based path planning methods.

## 7. Declaration of competing interest

The authors declare that they have no known competing financial interests or personal relationships that could have appeared to influence the work reported in this paper.

## 8. Data availability

The data will be provided upon request.

**AUTHORS**

**Tab. 3.** Comparison of Different Dynamic Path Planning Algorithms Across Different Scenarios

| Map Type | Case | Path Length | | | Time | | | Turning Radius | | |
|---|---|---|---|---|---|---|---|---|---|---|
| | | D-TIG | APF | APPATT | D-TIG | APF | APPATT | D-TIG | APF | APPATT |
| Short | C17 | 520.46 | 844.00 | 519.67 | 0.03 | 0.11 | 0.04 | 1.72 | 35.01 | 2.22 |
| | C18 | 485.87 | 718.00 | 491.07 | 0.03 | 0.08 | 0.03 | 0.64 | 147.98 | 3.45 |
| | C19 | 509.74 | 531.00 | 510.58 | 0.01 | 0.07 | 0.01 | 0.82 | 63.21 | 1.57 |
| | C20 | 625.68 | 885.00 | N/A | 0.04 | 0.12 | N/A | 3.08 | 13.03 | N/A |
| Large | C21 | 1010.41 | 1406.00 | 1010.01 | 0.07 | 0.14 | 0.05 | 1.38 | 10.65 | 1.90 |
| | C22 | 1125.06 | 1419.00 | 1124.14 | 0.06 | 0.14 | 0.04 | 1.99 | 9.66 | 2.28 |
| | C23 | 1036.91 | 1401.00 | 1033.27 | 0.04 | 0.14 | 0.03 | 3.52 | 1200.42 | 2.86 |
| | C24 | 990.64 | N/A | 994.53 | 0.03 | N/A | 0.04 | 1.95 | N/A | 3.40 |
| Sparse | C25 | 495.73 | 687.00 | 496.27 | 0.02 | 0.06 | 0.01 | 1.08 | 7.26 | 1.65 |
| | C26 | 516.49 | 532.00 | 518.80 | 0.01 | 0.04 | 0.01 | 1.08 | 2.98 | 2.17 |
| | C27 | 489.01 | 552.00 | 489.39 | 0.01 | 0.05 | 0.01 | 0.63 | 184.25 | 1.18 |
| | C28 | 508.54 | 536.00 | 513.59 | 0.01 | 0.05 | 0.01 | 2.21 | 8.04 | 3.59 |
| Dense | C29 | 558.54 | N/A | N/A | 0.05 | N/A | N/A | 6.51 | N/A | N/A |
| | C30 | 562.17 | N/A | N/A | 0.06 | N/A | N/A | 5.92 | N/A | N/A |
| | C31 | 505.38 | N/A | 505.14 | 0.04 | N/A | 0.05 | 1.86 | N/A | 2.59 |
| | C32 | 564.15 | N/A | N/A | 0.06 | N/A | N/A | 7.55 | N/A | N/A |

**Tab. 4.** Comparison of Different Dynamic Path Planning Algorithms Across Different Scenarios

| Map Type | Case | Path Length | | Time | | Turning Radius | |
|---|---|---|---|---|---|---|---|
| | | D-TIG | APPATT | D-TIG | APPATT | D-TIG | APPATT |
| Short | C17 | 520.46 | 571.56 | 0.02 | 0.03 | 0.55 | 4.12 |
| | C18 | 496.99 | 502.81 | 0.01 | 0.01 | 1.62 | 1.36 |
| | C19 | 513.46 | 520.09 | 0.005 | 0.007 | 0.97 | 1.47 |
| Large | C21 | 1017.12 | N/A | 0.02 | N/A | 2.03 | N/A |
| | C22 | 1130.68 | 1244.94 | 0.08 | 0.01 | 1.56 | 4.44 |
| | C23 | 1027.61 | 1044.94 | 0.03 | 0.01 | 2.70 | 3.99 |
| Sparse | C25 | 494.90 | 522.74 | 0.001 | 0.003 | 0.97 | 4.27 |
| | C26 | 534.93 | N/A | 0.002 | N/A | 2.2 | N/A |
| | C27 | 495.09 | 561.69 | 0.008 | 0.002 | 1.48 | 4.42 |

**Hichem Cheriet**[*] – Phd Student, SIMPA Laboratory, Université des Sciences et de la Technologie d'Oran, Bir El Djir 31000, Oran, Algeria, e-mail: hichem.cheriet@univ-usto.dz, www: N/A.

**Khellat Kihel Badra** – Department of Economics, Oran 2 Mohamed Ben Ahmed University, Bir El Djir 31000, Oran, Algeria, e-mail: khellat_badra@yahoo.fr, www: N/A.

**Chouraqui Samira** – Computer Science Department, Université des Sciences et de la Technologie d'Oran Mohamed Boudiaf, Bir El Djir 31000, Oran, Algeria, e-mail: samirachouraqui178@gmail.com, www: N/A.

[*]Corresponding author

## ACKNOWLEDGEMENTS

This work was supported by the research project "Modeling and Control of Aerial Manipulators" N° C00L07UN310220230004.